\PassOptionsToPackage{pdfpagelabels=false}{hyperref} 

\documentclass[letterpaper, 10 pt, conference]{ieeeconf}  % Comment this line out if you need a4paper
\IEEEoverridecommandlockouts                              % This command is only needed if 
\title{\LARGE \bf
Vision-based DRL Autonomous Driving Agent with Sim2Real Transfer}

\author{Dianzhao Li and Ostap Okhrin% <-this % stops a space
\thanks{Authors are with the Chair of Econometrics and Statistics, esp. in the Transport Sector, Technische Universität Dresden, Dresden, 01187, Germany and Center for Scalable Data Analytics and Artificial Intelligence (ScaDS.AI) Dresden/Leipzig, Germany.{ (E-mails: dianzhao.li@tu-dresden.de; ostap.okhrin@tu-dresden.de)}}%
}

\begin{document}

\maketitle
\thispagestyle{empty}
\pagestyle{empty}

%%%%%%%%%%%%%%%%%%%%%%%%%%%%%%%%%%%%%%%%%%%%%%%%%%%%%%%%%%%%%%%%%%%%%%%%%%%%%%%%
\begin{abstract}

To achieve fully autonomous driving, vehicles must be capable of continuously performing various driving tasks, including lane keeping and car following, both of which are fundamental and well-studied driving ones. However, previous studies have mainly focused on individual tasks, and car following tasks have typically relied on complete leader-follower information to attain optimal performance. To address this limitation, we propose a vision-based deep reinforcement learning (DRL) agent that can simultaneously perform lane keeping and car following maneuvers. To evaluate the performance of our DRL agent, we compare it with a baseline controller and use various performance metrics for quantitative analysis. Furthermore, we conduct a real-world evaluation to demonstrate the Sim2Real transfer capability of the trained DRL agent. To the best of our knowledge, our vision-based car following and lane keeping agent with Sim2Real transfer capability is the first of its kind. We have made the codes and the videos of the simulation and real-world evaluation accessible online\footnote{Code and videos are available on: \url{https://github.com/DailyL/Sim2Real_autonomous_vehicle}}.

\end{abstract}

%%%%%%%%%%%%%%%%%%%%%%%%%%%%%%%%%%%%%%%%%%%%%%%%%%%%%%%%%%%%%%%%%%%%%%%%%%%%%%%%
\section{INTRODUCTION}

In recent years, DRL has garnered significant attention from researchers as a powerful method to solve complex control problems across various fields, including Robotics [\citenum{kober2013reinforcement, andrychowicz2020learning, gu2017deep, haarnoja2017reinforcement}], Game playing [\citenum{silver2017mastering,mnih2015human,silver2016mastering,hessel2018rainbow,wurman2022outracing}], Finance \cite{jiang2017deep,yang2020deep}, and Natural language processing \cite{serban2017deep,zhang2019bridging}. In the field of autonomous driving, DRL has emerged as a natural tool to tackle the intricate decision-making required in complex situations. \par

The car following task is a basic driving task that can be addressed by rule-based control strategies, which are developed based on empirical observations of driver behavior and usually employ a set of fixed parameters to describe vehicle behavior. The Wiedemann car following model \cite{Wiedemann.1974} is one such model that describes the psycho-physiological aspects of driving behavior through four distinct driving regimes: free flow, approaching slower vehicles, car-following near the steady-state equilibrium, and critical situations requiring stronger braking actions. In addition to the Wiedemann car following model, intelligent driver model (IDM) \cite{treiber2000congested} is another widely used time-continuous car following model. It can generate realistic acceleration profiles and plausible behaviors in all single-lane traffic scenarios. IDM takes into account the desired speed, desired headway distance, and the difference between the actual and desired speed and headway. It can also model different driving behaviors such as aggressive and conservative driving. However, these models may not accurately capture the complex and dynamic behavior of drivers in real-world scenarios. To overcome this limitation, several studies have explored the use of DRL for the car following task. Considering the safety, efficiency, and comfort, the DRL car-following agent can provide improved performance in dynamic driving conditions [\citenum{lin2019longitudinal, yen2020proactive, naing2022dynamic}]. Furthermore, the real-world driving data can be used to train DRL agents for human-like behaviors \cite{zhu2018human} or improvement of driving policies \cite{li2023modified}. Besides utilizing various input features for vehicle control, vision-based car-following models exist that mimic the behavior of human drivers. These models rely on visual cues and processing to accurately track and follow other vehicles on the road, similar to how human drivers operate \cite{schwarzinger1992vision,zayed2004license}. Lane keeping is another essential component of autonomous driving that help to improve road safety and reduce driver fatigue. It uses a combination of sensors, cameras, and algorithms to detect lane markings on the road and provide feedback to the steering system and keep the vehicle centered within the lane. Lane keeping has been extensively researched with various solutions, for instance, PID controller, fuzzy logic controller, and DRL algorithm [\citenum{amditis2010situation, sallab2016end, basjaruddin2015lane}], etc.

In addition to training DRL agents in simulated environments, it is also a challenge to transfer the trained agents to real-world traffic environments (Sim2Real). Even though the simulators are getting realistic, there are still significant discrepancies between two environments (Sim2Real gap). To address this issue, several methods have been proposed, including domain randomization, domain adaptation, and knowledge distillation [\citenum{tobin2017domain, traore2019continual, hinton2015distilling}], to narrow the gap between the simulation and reality.
With the Sim2Real techniques, researchers are able to transfer the trained DRL agent to real-world to perform lane keeping, objective detection, overtaking, etc [\citenum{hu2022sim, reiher2020sim2real,muller2018driving,li2023platform}].\par

However, the aforementioned works are mainly focused on a single driving task such as lane keeping or car following, and the car following task mostly depends on multiple sensors to detect the gap between two vehicles and the velocity difference. To the best of our knowledge, there are no related works for vision-based multi-task driving agents with the abilities for Sim2Real transfer. This research gap motivated us to propose a robust DRL agent performing multi-task driving behavior that only depends on visual input. It is important to note that while this study may not fully replicate real-world conditions due to the specific environment setup, its primary objective is to showcase a promising direction for vision-based car following and lane keeping maneuvers. The study serves as a demonstration of the potential capabilities of the proposed approach, paving the way for further advancements and refinements to enhance the realism of the system and its applicability in practical scenarios. \par

\begin{figure*}[thpb]
      \centering
      \includegraphics[scale=0.75]{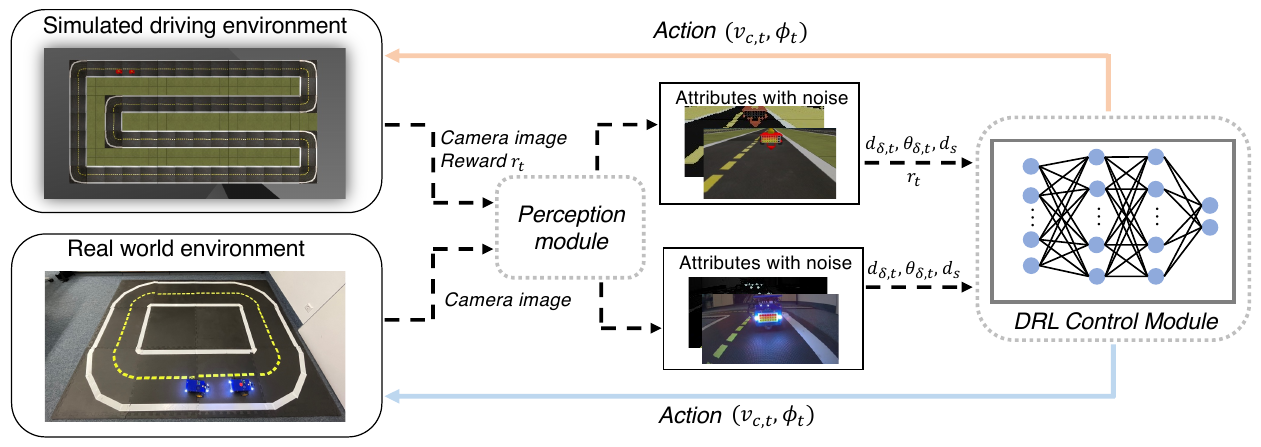}
      \caption{The proposed DRL framework for vision-based multi-task autonomous driving agents. The \emph{perception module} leverages camera images to produce impact attributes regarding the environment, then the DRL \emph{control module} utilizes the information to control the agent with enhanced generalization.}
      \label{fig:system}
   \end{figure*}

Summarizing, the main contributions of this work are as follows:

\begin{itemize}
    \item The training framework depicted in Fig. \ref{fig:system} with the separation of the \emph{perception module} and DRL \emph{control module} is proposed to narrow the Sim2Real gap. The inclusion of impact affordances derived from the \emph{perception module} has been demonstrated to enhance the stability of training and the overall generalization of the model.
    \item Considering safety, efficiency, and lane keeping capabilities, the trained DRL agent can perform stable car-following behavior, while no vehicle driving in front, the DRL agent shows excellent lane keeping ability.
    \item Compare with the baseline controller and evaluate the performance with multiple metrics, where the DRL agent achieves similar performance, even though it only depends on visual input.
    \item The Sim2Real transfer is performed for trained DRL agent in the real-world environment, and the generalization of the agent under the noise of the real world is evaluated.
\end{itemize}

\section{Background}

\subsection{Deep Reinforcement Learning}

Reinforcement Learning (RL) involves an agent interacting with an environment to maximize the reward it receives. This is done by treating the process as a Markov Decision Process (MDP), in which the agent observes a state $s_t$ of the environment at time step $t$, and then chooses an action $a_t$ according to the current policy $\pi(a_t|s_t)$. The environment transitions to a new state $s_{t+1}$ with probability $P(s_{t+1}|s_t, a_t)$, and provides a reward $r_{t+1}$ to the agent as feedback. The objective is to determine an optimal policy $\pi^*$ that maximizes the expected discounted cumulative rewards $\mathbb{E}_{\pi^*}\left\{ \sum_{k=0}^{\infty} {\gamma}^k r_{t+k+1} \right \}$, where $\gamma \in \left[0, 1\right]$ is a discount factor that determines the tradeoff between immediate and future rewards.\par

To measure the quality of a state $s_t$ for a given policy $\pi$, we use a state-value function $V^\pi$($s_t$) = $\mathbb{E}_\pi \left\{ R_t| s_t \right\}$, which represents the expected cumulative reward starting from state $s_t$ and following policy $\pi$. Similarly, the action-value function $Q^\pi$($s_t, a_t$) = $\mathbb{E}_\pi \left\{ R_t| s_t, a_t \right\}$ measures the expected cumulative reward starting from state $s_t$, taking action $a_t$, and then following policy $\pi$.\par

However, in practice, the state and action space can be vast and complex, making it computationally infeasible to use a tabular format to find optimal policies. To overcome this issue, DRL employs Deep Neural Networks to approximate the optimal Q-values and has shown remarkable success in solving complex tasks and achieving human-level performance.

\subsection{Proximal Policy Optimization (PPO)}

Proximal Policy Optimization (PPO) is a policy-based method, which means that it directly learns a policy, mapping from states to actions \cite{schulman2017proximal}. It uses a trust region optimization approach that restricts the size of the updates to the policy. This allows the algorithm to maintain the stability of the policy while still making progress toward the optimal policy. PPO uses a clipping parameter to limit the size of the policy updates, which prevents the new policy from diverging too far from the old policy and helps to ensure that the policy remains stable during training. Therefore, PPO is computationally efficient and scalable to handle large-scale problems. It is highly effective in a wide range of applications, including game playing, robotics, and natural language processing. In this study, we employed several DRL algorithms, implementing thorough hyperparameter tuning to explore their performance. After extensive experimentation, it was determined that the PPO algorithm consistently produced the most stable and robust training results.

\begin{algorithm}
\caption{PPO-Clip \cite{SpinningUp2018}}
\label{algo:ppo}
\begin{algorithmic}[1]
\State Input: initial policy parameter $\theta_0$ and value function parameter $\phi_0$.
\For {k = 0, 1, 2, ...}
\State Collect set of trajectories $\mathcal{D}_k = \{ \tau_i \}$ by running 
\NoNumber{policy $\pi_k = \pi(\theta_k)$ in the environment.}
\State {Compare rewards-to-go $\hat{R}_t$.}
\State Compute advantage estimates, $\hat{A}_t$ based on the 
\NoNumber{current value function $V_{\phi_k}$.}
\State Update the policy by maximizing the PPO-Clip \NoNumber{objective:}
\NoNumber{ \scalebox{0.65}{$\theta_{k+1} = \underset{\theta}{\mathrm{arg max}}\ \frac{1}{|\mathcal{D}_k|T} \sum\limits_{{\scriptscriptstyle \tau \in \mathcal{D}_k}}\sum\limits_{\scriptscriptstyle t=0}^ {\scriptscriptstyle T} \min\left[\frac{\pi_\theta(a_t|s_t)}{\pi_{\theta_k}(a_t|s_t)} A^{\pi_{\theta_k}}(s_t, a_t), g\left\{\epsilon, A^{\pi_{\theta_k}}(s_t, a_t)\right\}\right]$,}}
\NoNumber{via stochastic gradient ascent.}
\State Fit value function by regression on MSE:
{\par\centering 
    \scalebox{0.66}{ $\phi_{k+1} = \underset{\phi}{\mathrm{arg min}} \frac{1}{|\mathcal{D}_k|T} \sum\limits_{{\scriptscriptstyle \tau \in \mathcal{D}_k}}\sum\limits_{\scriptscriptstyle t=0}^ {\scriptscriptstyle T} {\left\{V_\phi(s_t) - \hat{R}_t\right\}}^2,$} \par}
\NoNumber{via gradient descent algorithm.}
\EndFor
\end{algorithmic}
\end{algorithm}

\subsection{Simulation and real-world platform}

In this study, we employ the Duckietown platform to train and assess the performance of our DRL agent. The platform as shown in Fig. \ref{fig:system} offers both simulation and real-world environments that enable researchers to train agents to execute diverse driving behaviors\cite{paull2017duckietown}.  The vehicle on this platform is equipped with a wide-angle monocular camera facing forward, which captures environmental information. To train and test autonomous driving agents using different methods, the Gym-Duckietown simulation is provided \cite{gym_duckietown}. It is developed using Python/OpenGL (Pyglet) and offers various customization options through different wrappers, such as one for the Robot Operating System (ROS), which facilitate the transition to the real robot. In the simulator, the agent is placed inside a Duckietown, which consists of a loop of roads with turns, intersections, obstacles, traffic lights, and other Duckiebots. \par

\begin{figure}[thpb]
      \centering
      \includegraphics[scale=0.45]{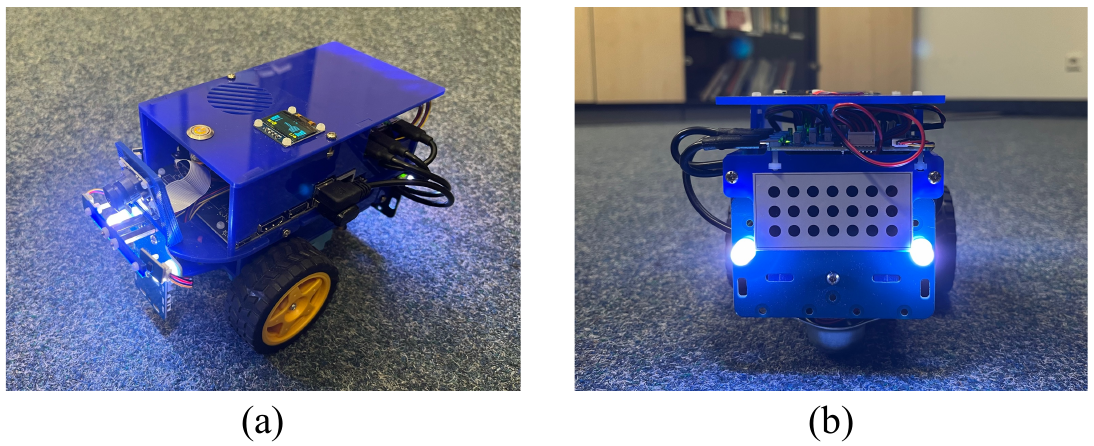}
      \caption{Robot car used during the real-world evaluation. (a) Side view of the robot car which equipped with front-view camera and Jetson Nano 2GB. (b) Back view of the car with a pattern of circles.}
      \label{fig:bot}
   \end{figure}

To evaluate the performance of the trained agent in the real world, we use the real-world Duckiebot shown in Fig. \ref{fig:bot}, which is equipped with a Jetson Nano 2GB computation unit responsible for image acquisition, processing, and vehicle control. Notably, the images obtained from the camera in the real-world environment may have visible differences compared to those obtained from the simulator, owing to variations in lighting conditions and camera angles. This presents a significant challenge when translating the trained agent into a real robot, which we have addressed in this study.

\begin{comment}
\begin{figure}
  \centering
  \begin{tabular}{ c @{\hspace{20pt}} c }
    \includegraphics[width=.45\columnwidth]{duckiebot_front.pdf} &
      \includegraphics[width=.45\columnwidth]{duckiebot_back.pdf} \\
    \small (a) &
      \small (b)
  \end{tabular}

  \medskip

  \caption{A figure caption}
\end{figure}
\end{comment}

\section{Multi-task driving model with DRL}

\subsection{Problem formulation}
\label{sec:problem}

Achieving fully autonomous driving requires addressing fundamental driving tasks such as car following, lane keeping, and overtaking. In this study, we focus specifically on lane keeping and car following tasks. Our autonomous agent is designed to use only camera images to detect lanes and sense the vehicle in front. When there is no vehicle ahead on the road, the agent must execute robust lane keeping behavior. When a slower-moving vehicle blocks its path, the agent must safely follow the vehicle while maintaining a safe distance and driving efficiency. To achieve a successful transfer of learned behavior from the simulated environment to the evaluation simulator and the real-world environment, the agent must be capable of accounting for any discrepancies between the simulation and the real-world scenarios. Moreover, the trained agent must consider the latency of image transfer and processing and control the vehicle drive properly with adequate tolerance. \par

\subsection{Perception module for image processing}
\label{sec:image processing}

Since the DRL agent only depends on images from the camera, the perception module is responsible for extracting task-related affordances. As shown in Fig. \ref{fig:system}, for our tasks, the perception module needs to detect the lanes on the road and the pattern on the back of the leading vehicle. \par

To enable lane detection, we follow a multi-step image processing pipeline \cite{paull2017duckietown}. This pipeline estimates two important parameters: the lateral displacement $d_\delta$ to the right lane center and the angle offset $\theta_\delta$ relative to the center axis. These estimates are later used in the control module to facilitate lane keeping. For the vision-based car following task, we attach a pattern of circles to the back of the leading vehicle as shown in Fig. \ref{fig:bot}. To detect the circles, we use the \emph{findCirclesGrid} function from OpenCV \cite{opencv_library}, which identifies the centers of the circle grid. Next, we compute the Euclidean distance $d_s$ between two circles to determine the distance between the two vehicles. As $d_s$ changes according to the distance between two vehicles, the change of  $d_s$ is used to indicate the time-to-collision between the two vehicles \cite{schwarzinger1992vision}.

\subsection{Action and observation space}
\label{sec:action}

This section covers the fundamental setup of our DRL agent. The action space of our autonomous vehicle is defined by the speed command $v_{c,t}$ and steering angle $\phi_t$, which follows the normal practice for such setups. To enable our DRL agent to perform lane keeping and car following maneuvers, we need to provide observations that support these tasks. As discussed before, the agent is divided into two distinct modules: the perception module and the control module to enhance the generalization. The perception module discussed in Section \ref{sec:image processing} is responsible for processing images captured by the front-view camera of the vehicle and extracting relevant environmental attributes required for driving. \par

For lane keeping task, the lateral deviation $d_{\delta}$ and orientation deviation $\theta_\delta$ processed from perception module are used. For vision-based car following task, to sense the velocity difference and the distance to leading vehicle, the distances between two points on the back of leading vehicle of current and last timestep are used. The overall input state $s_t$ for the DRL agent at time step $t$ is thus defined as 

\begin{equation}
    s_t =  {\left(\frac{d_{\delta, t}}{d_w}, \frac{\theta_{\delta, t}}{\pi},  v_t,  d_{s,t}, d_{s,t-1}, a_{v,t-1}, a_{\phi, t-1} \right)}^\top ,
    \label{eq:input state}
    \notag
\end{equation}
where $d_w$ indicates the lane width, $v_t$ denotes the velocity of the ego vehicle, $d_{s,t}$ and $d_{s,t-1}$ are the distances between two points on the back of leading vehicle of current and last timestep. Additionally, action output from control module $a_{v,t-1}$ and $a_{\phi, t-1}$ of last timestep are used to assist the agent to perform more smooth control. Here we use the work from \cite{li2023platform} as a source of reference for lane keeping task, but this work is mainly focused on the combination of car following and lane keeping task, as well as its evaluation.

\subsection{Reward function}

The reward function is the cornerstone of any RL algorithm, as it guides the behavior of the trained agent. To achieve the desired outcome, it is crucial to consider the lane keeping and car following capabilities of the agent. To address this, we design a reward function with two primary sub-rewards: one for lane keeping and another for car following.  To ensure the ability of the agent to accurately follow the lane, we utilize two key metrics: the lateral deviation and orientation deviation relative to the right lane center. These measures allow us to closely monitor the position and orientation of the vehicle, and ensure that it stays within the designated lane with only a small margin of error. To achieve optimal car following behavior, we take into account two key factors: safety and driving efficiency. By carefully balancing these elements, we can ensure that the agent maintains a safe following distance while also driving efficiently.

\subsubsection{Safety}
Ensuring safety is the top priority, the agent needs to maintain a safe distance from the vehicle in front to avoid any potential collisions. To support this, we use a metric known as time-to-collision (TTC), which allows us to penalize any unsafe driving behavior. TTC is a widely used safety factor that indicates the amount of time left before two vehicles collide, based on their current velocities:
    \begin{equation}
        \text{TTC} (t) = \frac{d_{n-1,n}(t)}{\Delta v_{n-1,n}(t)},
        \notag
    \end{equation}
    where $d_{n-1,n}(t)$ is the bumper-to-bumper diastance between leader $n-1$ and follower $n$, $\Delta v_{n-1,n}(t)$ represent the speed difference between two vehicles.
    To incorporate TTC as safety subreward, a lower bound value of TTC should be determined. Considering that the vehicle used in our simulation and real-world is miniature model, thus we choose the TTC lower bound as 1.5s. Then we construct the  safety subreward $R_{safe}$ as: 
    
    \begin{equation}
        R_{safe}(\text{TTC}) =
        \begin{cases} 
        \log(\frac{\text{TTC}}{1.5}), & 0 < \text{TTC} \le 1.5s,\\
        0, & \text{otherwise}.
        \end{cases}
        \notag
    \end{equation}
    In this way, if the TTC is less than 1.5 seconds, the agent will receive a penalty, and as the TTC approaches zero, the reward for safety $R_{safe}$ will become increasingly negative, potentially approaching negative infinity.
\subsubsection{Driving efficiency}
\label{sec:headway}
To quantity measure the driving efficiency, we employ time headway, which represents the time for the ego vehicle to reach the back of the leading vehicle. 
    \begin{equation}
        h (t) = \frac{d_{n-1,n}(t) + l_{n-1}}{v_{n}(t)},
        \notag
    \end{equation}
    where $l_{n-1}$ is the length of leading vehicle $n-1$ and $v_{n}(t)$ is the speed of ego vehicle. To utilize time headway in the reward function, we construct it with log-normal distribution $H \sim \text{Lognormal}(\mu, \delta^2)$. Next, the driving efficiency reward component $R_{eff}$ is showed with probability density function (pdf) for time headway:
    
    \begin{equation}
        R_{eff}(h) = \frac{1}{\sqrt{2\pi}h\delta} \exp\left\{-\frac{(\ln(h)-\mu)^2}{2\delta^2}\right\}.
        \notag
    \end{equation}
    We choose $\mu = 0.4226$ and $\delta = 0.4365$ based on empirical data \cite{zhu2020safe}. The reward for time headway is maximized when it is close to 1.26s, and a smaller or larger headway results in a smaller reward value.

\subsubsection{Lateral and orientation deviation}

For Lane keeping ability of the agent, we should consider the lateral and orientation deviation relative to the right lane center. Therefore, we incorporate a penalty for the lateral deviation by measuring the cross-track error ($e_y(t)$) towards the target path with the following equation:

\begin{equation}
    R_{l,\delta}(t)= a^{k_{r_{c}} e_y(t)} - 1,
    \notag
\end{equation}
where $a$ and $k_{r_{c}}$ are tunable parameters that controls sensitivity and set as 0.001 and 0.6 respectively. By employing this equation, we ensure that the agent is not penalized when the cross-track error is precisely zero, thus encouraging it to maintain accurate alignment with the desired trajectory. As the magnitude of $e_y(t)$ grows larger, the reward gracefully diminishes, asymptotically approaching a value of -1. This guides the behavior of the agent by encouraging it to minimize cross-track errors while navigating the environment.\par

For orientation deviation penalty, the agent uses the angle offset $\chi_{yaw,t}$ between the current heading and the desired heading: 

\begin{equation}
    R_{\phi,\delta}(t)= \lvert\chi_{yaw,t}\rvert.
    \notag
\end{equation}

Therefore, the overall reward function is defined as the sum of these reward components. As the weights measure the trade-off of different components, we conduct extensive experiments to test different weight combinations. Through this rigorous testing process, we identify and select the set of weights that produces the best overall performance.

\begin{equation}
    R = w_{s} R_{safe} + w_{e} R_{eff} + w_{l,\delta} R_{l,\delta} + w_{\phi,\delta} R_{\phi,\delta},
    \notag
\end{equation}
where $w_{s}$ and $w_{e}$ are set as 0.5 and 0.8 respectively and the car following weights $w_{l,\delta}$ and $w_{\phi,\delta}$ are set as 0.3 and -0.2.

\section{Experiments and results}

\subsection{Training process}

\begin{figure}[thpb]
      \centering
      \includegraphics[width=\columnwidth]{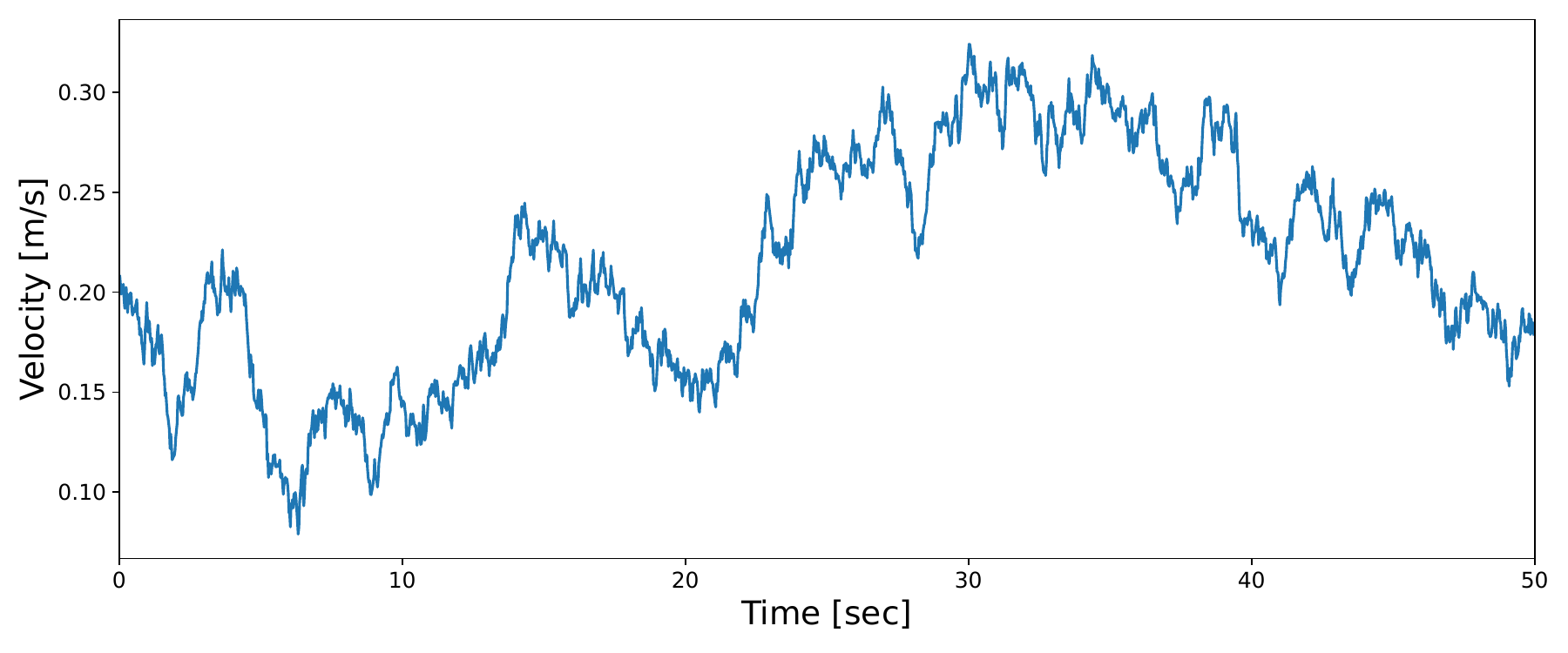}
      \caption{An example velocity trajectory for the leading vehicle generated by the Ornstein-Uhlenbeck process for the training process.}
      \label{fig:leader_speed}
   \end{figure}

To train multi-task agents for autonomous driving, we need to create a leader-follower pair. Therefore, we use the Ornstein-Uhlenbeck process to generate random leader trajectories \cite{uhlenbeck1930theory}, which helps the agent learn to follow a variety of driving patterns and improves their ability to generalize to new situations. Fig. \ref{fig:leader_speed} shows an example leader velocity trajectory generated using the Ornstein-Uhlenbeck (OU) process. Before each episode of the training process, random velocity trajectories will be generated and control the leading robot to run with the desired velocity. For steering angle control, the leader uses the proportional–derivative (PD) controller baseline provided by Gym-Duckietown \cite{gym_duckietown}. Therefore, the leader can perform lane keeping behavior with desired random velocity. The DRL agent must effectively follow the leading vehicle by maintaining a safe distance and suitable velocity, all while executing proper lane keeping behavior. The initial bumper-to-bumper gap between two vehicles is set randomly in the range [\SI{0}{m}, \SI{2.0}{m}]. \par

\begin{figure}[thpb]
      \centering
      \includegraphics[width=\columnwidth]{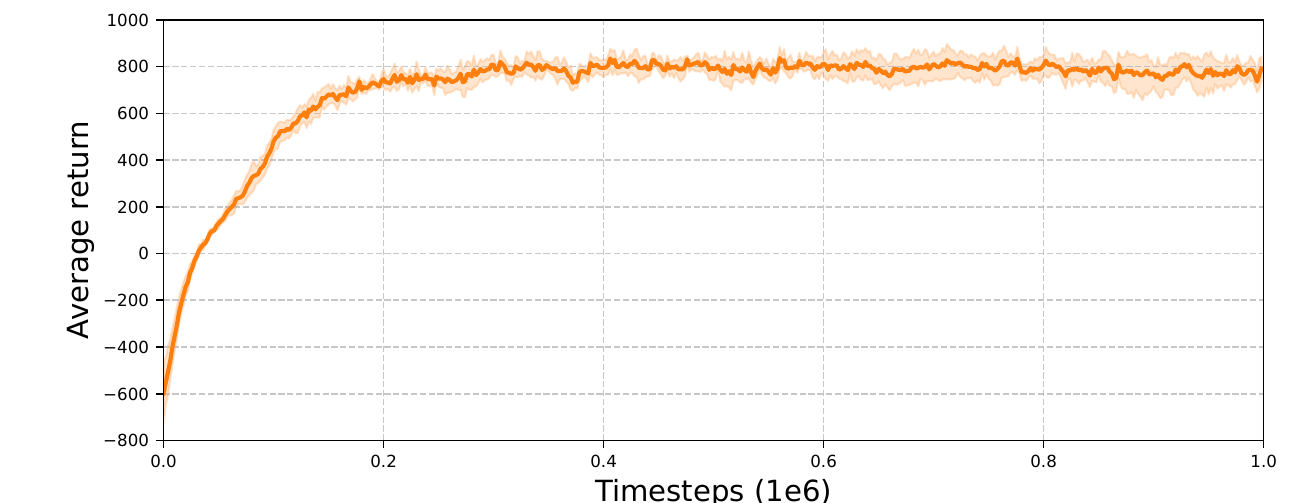}
      \caption{Training results of PPO agent with ten independent seeds over one million steps.}
      \label{fig:reward}
   \end{figure}

The training process utilizes an NVIDIA GeForce RTX 3080 GPU and typically requires approximately seven hours to complete one million steps of interaction with the running frequency of 40 Hz in the simulated environment. The training is initiated with ten varying seeds and executed for one million timesteps on multiple occasions with the hyperparameters in Table \ref{tab:ppo_parameter}. The successful convergence of the PPO agent during the training process is depicted in Fig. \ref{fig:reward}, leading to a stable and consistent performance at the end of the training. \par

\begin{table}[ht]
\caption{Hyperparameters used for PPO during training process.}
\label{tab:ppo_parameter}
\begin{center}
\begin{tabular}{ll}
\toprule
Hyperparameters  & Value\\
\midrule
 Neural network structure & 2 $\times$ $\left[ 64, \rm{ReLU} \right]$ \\
 Learning rate ($l_r$)  & $5\times10^{-4}$\\
 Reward discount factor ($\gamma$) & 0.99\\
 Value function coefficient ($c_1$) & 0.5\\
 Clip range ($\epsilon$)  &0.2 \\
 GAE parameter ($\lambda$) & 0.95\\
 Number of SGD epochs & 10\\
 Minibatch size &  64\\

\bottomrule
\end{tabular}
\end{center}
\end{table}

\subsection{Evaluation in simulation}

Next, we evaluate our model in a simulation using the Gym-Duckietown environment. Since there are no existing benchmarks for vision-based lane keeping and car following models, we compare our approach to a combination of IDM and PD controllers. The IDM controller uses information such as speed difference to the leading vehicle and bumper-to-bumper gap to regulate the speed of the ego vehicle and achieve car following behavior. Meanwhile, the PD controller provided by Gym-Duckietown controls the steering angle to perform lane keeping maneuvers. Several experiments are conducted to determine the best parameters for the baseline agent, which are presented in Table \ref{tab:baseline_parameter}.    

\begin{table}[ht]
\caption{Parameters used for baseline controller in the evaluation.}
\label{tab:baseline_parameter}
\begin{center}
\begin{tabular}{cll}
\toprule
Symbol  & Description   & Value\\
\midrule
$v_{des}$  & Desired velocity & \SI{1.0}{m/s}\\
		$T$ & Safe time gap & \SI{1}{s}\\
		$a$  & Maximum acceleration & \SI{1}{m/s^2} \\
		$b_{comf}$  & Comfortable deceleration & \SI{1}{m/s^2}\\
		$g_{min}$  & Minimum gap   & \SI{0.2}{m}\\
		$K_p$ & Proportional gain & 2.0\\
		$K_d$ & Derivative gain & 5.0\\
\bottomrule
\end{tabular}
\end{center}
\end{table}

\begin{figure}[thpb]
      \centering
      \includegraphics[scale=0.4]{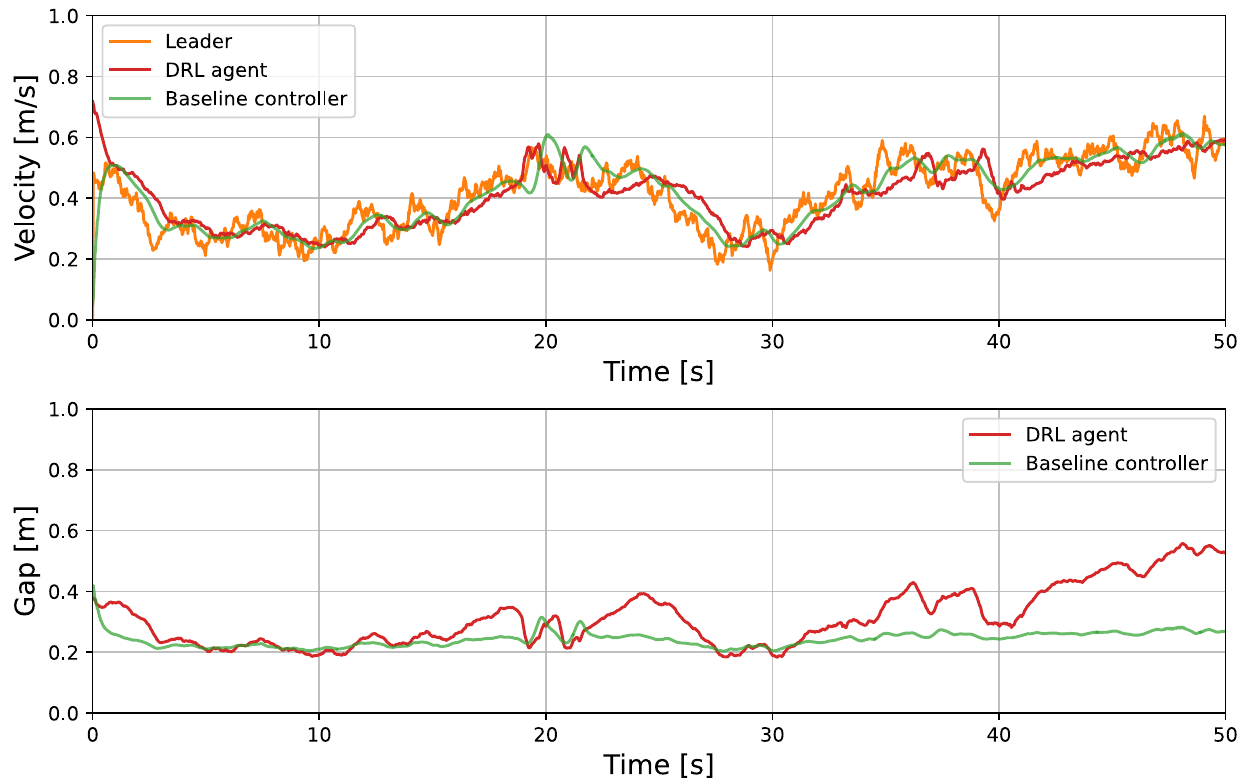}
      \caption{Example trajectories of DRL agent (red) and baseline controller (green) following random leader trajectory (orange) during the evaluation.}
      \label{fig:evaluation in simulation with random}
   \end{figure}

\begin{figure}[thpb]
      \centering
      \includegraphics[scale=0.4]{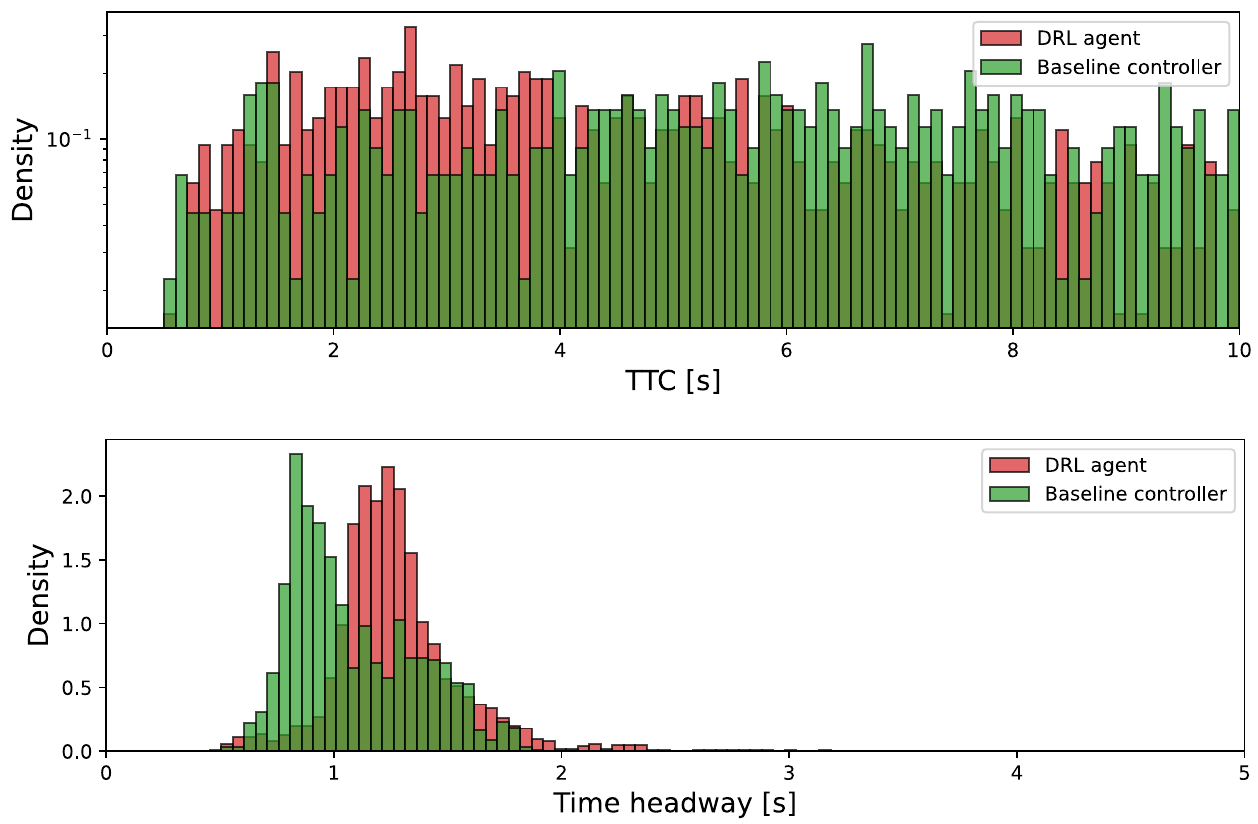}
      \caption{Distribution of TTC and time headway for DRL agent (red) and baseline controller (green).}
      \label{fig:ttc and headway evaluation in simulation random}
   \end{figure}

First, we assess the car following performance for both the DRL agent and the baseline agent using distinct sets of random leader velocity trajectories, which are generated via the OU process explained during the training phase. Fig. \ref{fig:evaluation in simulation with random} depicts the results of the evaluation, where both agents follow the leader with randomly generated velocities. As shown in Fig. \ref{fig:evaluation in simulation with random}, the DRL agent and IDM controller are following close to the leading vehicle, with IDM showing a more aggressive driving behavior and maintaining smaller distances. For a more detailed analysis, we compare the TTC and time headway of both followers, where the TTC is less than 10 seconds and the time headway is less than 5 seconds. As depicted in Fig. \ref{fig:ttc and headway evaluation in simulation random} and Table \ref{tab:evaluation results random}, despite the baseline controller has access to all the necessary information regarding the car following and lane keeping maneuvers, the vision-based DRL agent still achieve similar performance. We consider a TTC below 0.5 seconds as a safety-critical situation for the miniature vehicle. DRL agent has a minimal value of 0.53 seconds and for baseline controller is 0.56 seconds. Two followers are both performing safe car following behaviors. From Fig. \ref{fig:ttc and headway evaluation in simulation random}, we also observe that the time headway of the DRL agent is mainly distributed around 1.26 seconds, which is the optimal value according to the setup of the reward function discussed in Section \ref{sec:headway}. \par

During the validation process, the lane keeping performance for both agents is also assessed using four metrics: survival time ($T_s$), lateral deviation ($\delta_l$) from the center of the right lane, orientation deviation ($\delta_\phi$) from the tangent of the right lane center, and major infractions ($i_m$) which measure the amount of time the vehicle spends outside of the drivable zones. The survival time ($T_s$) metric indicates the length of time until the simulation is terminated, which occurs when the vehicle drives outside of the road. The lateral deviation ($\delta_l$) and orientation deviation ($\delta_\phi$) are both integrated over time to assess the overall performance. The major infractions ($i_m$) metric measures the amount of time the vehicle spends outside of the drivable zones. All four metrics are computed as the median value of 10 episodes, each lasting for 50 seconds. Table \ref{tab:evaluation results random} shows that the DRL agent achieved a similar level of performance compared to the baseline agent, which was controlled by a PD controller and had full access to the true information of the environment. Remarkably, the DRL agent performed well despite not having access to the true information of the environment. Neither controller had any major infractions, indicating that both were able to maintain good lane-keeping ability and avoid crossing the middle line while following the leading vehicle. Moreover, the DRL agent achieved a smaller lateral deviation during evaluation.\par

\begin{table}[ht]
    \centering
    \caption{Evaluation results of two agents following random generated leading vehicle trajectories.}
    \label{tab:evaluation results random}
    \resizebox{\columnwidth}{!}{
    \begin{tabular}{llcc}
    \toprule
    \multicolumn{2}{l}{Performance metrics} & DRL agent   & Baseline controller\\
    \midrule
    \multirow{4}{*}{TTC [s]} & Min. & 0.53 & 0.56\\
    & Mean & 4.47 & 5.58\\
    & Med. & 3.93 & 5.79\\
    & Std. & 2.39 & 2.50\\
    \hline
    \multirowcell{4}[0pt][l]{Time\\headway [s]} & Min. & 0.53 & 0.49\\
    & Mean & 1.27 & 1.08\\
    & Med. & 1.24 & 1.00\\
    & Std. & 0.28 & 0.28\\
    \hline
    \multicolumn{2}{l}{Survival Time ($T_s$) [s]} & 50 & 50\\
    \multicolumn{2}{l}{Lateral deviation ($\delta_l$) [m$\cdot$s]} & 1.10 & 1.11 \\
    \multicolumn{2}{l}{Orient. deviation ($\delta_\phi$) [rad$\cdot$s]} & 2.81 & 0.87\\
    \multicolumn{2}{l}{Major infractions ($i_m$) [s]} & 0.0 & 0.0\\
	\bottomrule
    \end{tabular}}
\end{table}

\begin{figure}[thpb]
      \centering
      \includegraphics[scale=0.41]{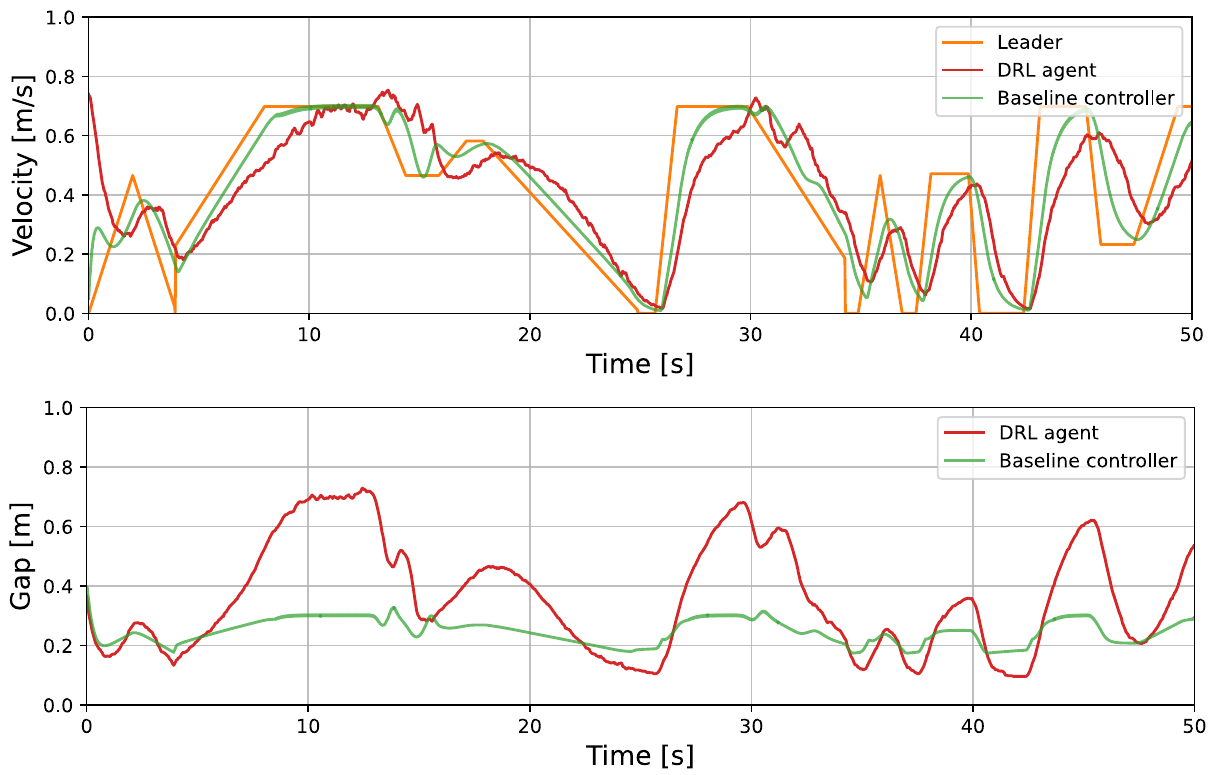}
      \caption{Example trajectories of DRL agent (red) and baseline controller (green) following a self defined external leader trajectory (orange) during the evaluation.}
      \label{fig:evaluation in simulation with self defined}
   \end{figure}

For a more comprehensive evaluation,  a leading vehicle speed profile is designed that takes into account safety-critical driving situations, which do not appear in normal driving scenarios. Fig. \ref{fig:evaluation in simulation with self defined} illustrates the vehicle trajectories of the DRL agent and baseline agent following the leading vehicle with the self-defined speed profile. As the leader starts with normal acceleration and reaches the velocity of \SI{0.7}{m/s}, both the DRL agent and baseline controller follow this change. The IDM controller is more aggressive and follows the leader closer. However, the DRL agent, which obtained images to follow the leader, exhibits a more passive driving style and maintained a greater distance from the leader. From $T=\SI{13}{s}$, the leader performs non-safety-critical braking until it stops. The DRL agent and baseline controller both decelerate properly and keep suitable distances from the leader. Later, from $T=\SI{33}{s}$, the leader performs several safety-critical accelerating and braking maneuvers. We observe that both controllers handle these situations well and maintained safe distances from the leader.\par

\begin{figure}[thpb]
      \centering
      \includegraphics[scale=0.41]{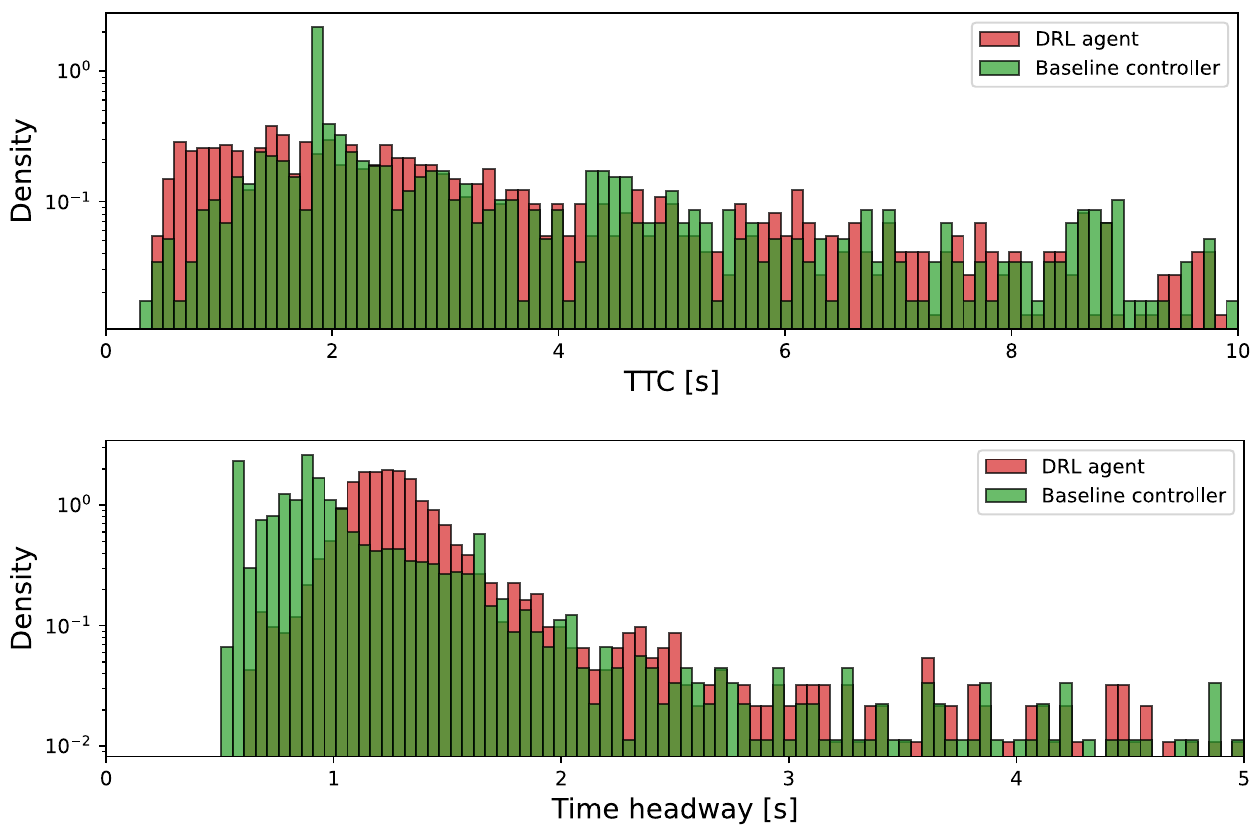}
      \caption{Distribution of TTC and time headway for DRL agent (red) and baseline controller (green).}
      \label{fig:ttc and headway evaluation in simulation}
   \end{figure}

The TTC and time headway of both followers at each time step are displayed in Fig. \ref{fig:ttc and headway evaluation in simulation}. According to Table \ref{tab:evaluation results self defined}, the DRL agent has a lower TTC bound of 0.48 seconds, which is still within the tolerance level, despite the unusual braking behavior of the leading vehicle during the evaluation. In contrast, the IDM agent has a minimal value of 0.38 seconds, which falls under the category of serious conditions. Moreover, the TTC distributions of both followers are mainly concentrated around 3s with similar variance considering TTC under 10s. Regarding the time headway, both followers exhibit similar performance, with the values of DRL mainly distributed around 1.2 seconds. \par

Considering the lane keeping performance under this safety-critical circumstance, Table \ref{tab:evaluation results self defined} indicates that the DRL agent and baseline controller are still able to handle the lane keeping maneuver while maintaining suitable car following capabilities. The lateral deviation of the DRL agent is smaller than the baseline controller and both agents have zero major infractions.  

\begin{table}[ht]
    \centering
    \caption{Evaluation results of two agents following self-defined leading vehicle trajectories.}
    \label{tab:evaluation results self defined}
    \resizebox{\columnwidth}{!}{
    \begin{tabular}{llcc}
    \toprule
    \multicolumn{2}{l}{Performance metrics} & DRL agent   & Baseline controller\\
    \midrule
    \multirow{4}{*}{TTC [s]} & Min. & 0.48 & 0.38\\
    & Mean & 3.25 & 3.39\\
    & Med. & 2.54 & 2.32\\
    & Std. & 2.28 & 2.26\\
    \hline
    \multirowcell{4}[0pt][l]{Time\\headway [s]} & Min. & 0.62 & 0.54\\
    & Mean & 1.40 & 1.12\\
    & Med. & 1.27 & 0.91\\
    & Std. & 0.55 & 0.63\\
    \hline
    \multicolumn{2}{l}{Survival Time ($T_s$) [s]} & 50 & 50\\
    \multicolumn{2}{l}{Lateral deviation ($\delta_l$) [m$\cdot$s]} & 1.09 &  1.12\\
    \multicolumn{2}{l}{Orient. deviation ($\delta_\phi$) [rad$\cdot$s]} & 3.51 & 0.96\\
    \multicolumn{2}{l}{Major infractions ($i_m$) [s]} & 0.0 & 0.0\\
	\bottomrule
    \end{tabular}}
\end{table}

\subsection{Evaluation in real-world environment}

For the Sim2Real transformation, the real-world scenario is used to assess the lane keeping and car following abilities of the trained DRL agent in the real world. However, due to the unavailability of accurate information about other vehicles, IDM cannot be applied in the real world. Consequently, our evaluation focuses solely on demonstrating the car following ability of the DRL agent. During the evaluation process, we utilize a PID baseline to control one leading vehicle, which demonstrates lane keeping behavior along the track. Simultaneously, the ego vehicle, controlled by the DRL agent, aims to perform both car following and lane keeping tasks. The real-world evaluation is conducted at a running frequency of 40 Hz, mirroring the setup used in the simulation. This consistency ensures that the evaluation process remains comparable to the simulated environment, enabling accurate and reliable assessments of the performance in real-world scenarios. As demonstrated in the evaluation video, the DRL agent effectively guides the ego vehicle to follow the leader at an appropriate speed. When the leading vehicle comes to a stop, the ego vehicle also halts after it, maintaining a suitable distance.
The evaluation videos are available on the project website.

\section{CONCLUSIONS}

This study proposes a vision-based DRL agent that can simultaneously perform lane keeping and car following tasks. The overall system is divided into two modules: the perception module and the control module. The perception module extracts task-relevant attributes of the surroundings, while the control module is a DRL agent that takes these attributes as input. To evaluate the performance of the DRL agent, we compare it with a baseline algorithm in both simulation and real-world environments. In the simulation, we compare the car following and lane keeping capabilities of the DRL agent and baseline controller using various performance metrics. In the real-world environment, we demonstrate that the DRL agent can follow the leading vehicle while maintaining lane keeping ability. In future work, we plan to enhance our DRL agent by incorporating a comfort factor to address unstable driving behavior. Additionally, we aim to deploy more advanced algorithms for improved generalization.

\addtolength{\textheight}{-0.5cm}   % This command serves to balance the column lengths

%%%%%%%%%%%%%%%%%%%%%%%%%%%%%%%%%%%%%%%%%%%%%%%%%%%%%%%%%%%%%%%%%%%%%%%%%%%%%%%%

%%%%%%%%%%%%%%%%%%%%%%%%%%%%%%%%%%%%%%%%%%%%%%%%%%%%%%%%%%%%%%%%%%%%%%%%%%%%%%%%

%%%%%%%%%%%%%%%%%%%%%%%%%%%%%%%%%%%%%%%%%%%%%%%%%%%%%%%%%%%%%%%%%%%%%%%%%%%%%%%%

\section*{ACKNOWLEDGMENT}

This work was funded by Center for Scalable Data Analytics and Artificial Intelligence (ScaDS.AI) Dresden/Leipzig, Germany. The authors would like to thank Fabian Hart, Martin Waltz, Niklas Paulig, and Ankit Chaudhari for constructive criticism of the manuscript, and also thank Martin Treiber, for providing the IDM parameters for evaluation.

\bibliographystyle{IEEEtran}
\bibliography{references}{}

\end{document}